\documentclass{article}

\usepackage{arxiv}

\usepackage[utf8]{inputenc} 
\usepackage[T1]{fontenc}    
\usepackage{hyperref}       
\usepackage{url}            
\usepackage{booktabs}       
\usepackage{amsfonts}       
\usepackage{nicefrac}       
\usepackage{microtype}      
\usepackage{lipsum}
\usepackage{graphicx}
\graphicspath{ {./images/} }
\usepackage{graphicx}

\usepackage[square]{natbib}
\renewcommand\cite{\citep}  

\usepackage[normalem]{ulem}
\usepackage{xcolor}

\def\ODdel#1{\bgroup\markoverwith{\textcolor{cyan!80!yellow!80!black}{\rule[0.5ex]{2pt}{1pt}}}\ULon{#1}}

\title{End to End Dialogue Transformer}

\author{
 Ondřej Měkota \\
  \texttt{ondrej.mekota@me.com} \\
  \AND
  Memduh Gökırmak \\
  \texttt{gokirmak@seznam.cz}
  \AND
  Petr Laitoch \\
  \texttt{petrlaitoch@yahoo.com}
}

\begin{document}
\maketitle
\begin{abstract}
Dialogue systems attempt to facilitate conversations between humans and computers,
for purposes as diverse as small talk to booking a vacation.  
We are here inspired by the performance of the recurrent neural 
network-based model Sequicity \cite{lei}, which when conducting a dialogue
uses a sequence-to-sequence architecture to first produce a textual
representation of what is going on in the dialogue, and in a further step 
use this along with database findings to produce a reply to the user.
We here propose a dialogue system based on the Transformer architecture 
instead of Sequicity's RNN-based architecture, that works similarly
in an end-to-end, sequence-to-sequence fashion.  
\end{abstract}


\section{Introduction}
Dialogue systems aim to simulate a conversation between a user and a computer. These systems
can be built simply for the purpose of chatting, but are often aimed at helping a user 
complete some task or retrieve some information without the need for interacting with
another human \cite{wen2016networkbased, wen-etal-2016-conditional}.

One the challenges of dialogue systems is to both provide the information the user requires 
and to present it in the form of fluently generated responses. Sequicity \cite{lei} aims to 
solve this issue by decoding the response in two steps: first explicitly decoding a belief state and then conditioning the final response on it.
Sequicity also presents an \emph{end to end} solution to dialogue systems, where the steps
of understanding the user's intention, keeping track of what's going on in the conversation
and producing an intelligible and correct response are all handled by the same model.

The Transformer \cite{vaswani} architecture has significantly improved  the benchmark in
machine translation and in recent years it has been used as a language model in various NLP applications.
Inspired by Sequicity, we here attempt to use a Transformer model to implement 
a sequence to sequence neural dialogue system that works end to end.

\section{Related works}
Sequicity
\cite{lei} is one of the first popular end-to-end dialog system architectures. It uses an RNN-based sequence-to-sequence (seq2seq) \cite{seq2seq} model further enhanced by the Two-Stage
Copy Net inspired by \cite{gu}.  It is design to tackle previous modular pipeline
designs. It uses an annotated belief state to accomplish separate dialog
tracking.

In recent years, the Transformer \cite{vaswani}, initially introduced in the
machine translation, has become the dominant neural sequence to sequence
architecture improving upon convolutional and recurrent \cite{seq2seq} networks. One of the potential downsides of the Transformer is that it
contains more parameters and requires more training epochs and a larger
training dataset compared to other models. However, the model also allows for
increased parallel computation and reduces time to convergence.

Large pre-trained models \cite{devlin,radford,t5,gpt3} based on the Transformer
architecture are becoming extremely popular. They aim to reduce the amount of
necessary domain-specific training data by pre-training on vast amounts of
general text data.

Newer end-to-end dialog system architectures such as those of \citet{budzianowski} or
\citet{WuZhang} make use of the pre-trained GPT-2 model.  They aim to prove that large
generative models pre-trained on large general-domain corpora can support
task-oriented dialogue applications. \citet{budzianowski}'system is based on the
TransferTransfo framework and is fine-tuned to a dialog dataset.  The basic
idea behind the Alternating Roles Dialog Model (ARDM) \cite{WuZhang} is to simultaneously model the user and system
with two separate GPT-2s to capture the different language styles. ARDM even
requires no human supervision such as belief states or dialog acts.

\section{Transformer E2E}

\begin{figure}
    \centering
    \includegraphics[width=0.6\textwidth]{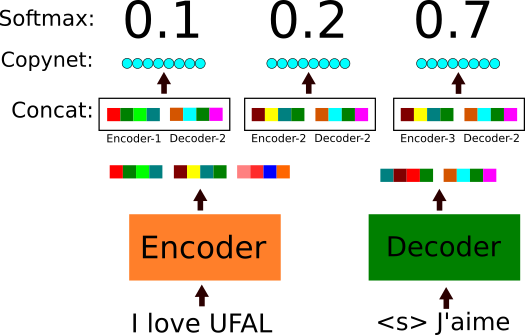}
    \caption{Producing a distribution over the encoder tokens when producing the decoder
    output at its second position. This component
    decides not \emph{whether} to copy, but the likelihood of each encoder
    input token being copied over, if we were to copy one of them. The example
    here is translation, of the English sentence \emph{I love UFAL} to
    French \emph{J'aime UFAL}. We want UFAL to be the token most likely to
    be copied, and train our copynet to copy the appropriate token.}
    \label{fig:copynet}
\end{figure}
The Transformer \cite{vaswani} is an encoder-decoder neural architecture that 
uses self-attention to take advantage of the contextual
information provided by the other tokens while processing each token. 
This architecture provides an efficient and parallelizable use of 
an attention mechanism and also establishes the new state of the art in
a number of tasks including machine translation in several language
pairs.

Our system, based on the Transformer architecture, deals with input from a user in two steps.
In the first step, the user input,  previous dialogue state and previous system response
are taken and processed to produce the new dialogue state. In the second step, along with the original
three inputs, the results of a database lookup and the newly decoded dialogue state are used. All
of these together are passed through the system in the second step to produce the response
of the system in the current stage of the dialogue.

Our model is distinct from the Transformer architecture in that it has a single 
encoder but two decoders. The same encoder encodes the input, and this encoding is used both for updating the 
dialogue state (bspan) and generating the response. A separate bspan decoder and a 
response decoder are used for these respective tasks when processing input from the encoder.

We implement a mechanism similar to Copynet \cite{gu} as used in the original Sequicity implementation.
Both of the decoders produce a distribution over the target vocabulary of words to be generated,
but also a probability distribution over the words input to the encoder. 
A value $P_{\mathrm{gen}}$, the probability of generating a new word versus copying a given output token at each step, is also determined at each time step.

Our copy mechanism takes the final encoded representations of each input
position from the encoder, and concatenates them with the current decoder output, to estimate the probability of copying 
over the token at any input position to the current output position. 
The result of this
concatenation is given to a dense feed-forward network that we call our
``copy network" to produce this probability. The values corresponding to
each input position are soft-maxed to produce a 
different copy
distribution over the encoder tokens for each decoder token. At each step of decoding,
the probabilities of generating or copying each token in the vocabulary are weighted by $P_{\mathrm{gen}}$
and $1 - P_{\mathrm{gen}}$ respectively, and added together. So we separately estimate
the likelihood of each input token to be copied to any output position, 
and whether this copying should take place at all.

\section{Evaluation}
\label{sec:eval}

We compare our system with \cite{lei} on the Camrest676 \cite{camrest} dataset. 
It contains $408$ training dialogues, and $136$ test and development dialogues within the domain of restaurant reservation. 
It has been delexicalized: each slot (restaurant name, food type, etc.) has been replaced by a generic token ("name\_SLOT", "food\_SLOT", etc.).
There are, on average, $4$ turns per dialogue.
The size of the vocabulary is $811$ for user queries and $1470$ for system responses.

We performed a grid search on the development part of the data over the dimensionality of the feed-forward layer ($256$, $512$, $1024$), the number of attention heads ($1$, $2$, $5$, $10$), the number of hidden layers ($2$, $3$, $4$), dropout rate ($0.1$, $0.2$), and the number of warmup steps ($100$, $500$, $4000$, $6000$, $12000$). 
Then we selected the several best models and further tuned the word embedding dimensionality ($50$, $100$)

Our best performing configuration of the Transformer uses $3$ layers, 
$1$ attention head, $512$ dimensions in the feed-forward layers
and a dropout \cite{dropout} rate of $0.1$.
We use $50$-dim GloVe \cite{glove} word embeddings as initial parameters and train our network in mini-batches of size $32$ for $54$ epochs.

We evaluate the systems using ``success F1 score'' which measures the F1 score of matched slots (food\_SLOT, name\_SLOT, etc.) in all system responses for all test set dialogues, matching against reference responses, and also by the BLEU \cite{bleu} score of the generated responses. 
BLEU measures how many word n-grams are shared with the gold data, for $n=1,2,3,4$ (with some penalization for short outputs).

\begin{table}[h!]
\centering
\begin{tabular}{p{5cm}rr}
\toprule
System                                  & Success F1        & BLEU    \\
\midrule
Transformer (no Copynet)                & 0.710             & 0.315    \\
Transformer + Copynet                        & 0.770             & \textbf{0.327}    \\
Sequicity                             & \textbf{0.854}              & 0.253 \\
\bottomrule
\end{tabular}

\caption{Comparison of dialogue systems described in Section~\ref{sec:eval} in terms of success rate and BLEU.}\label{tab01}
\end{table}

We compare above our Transformer model both without and with the Copynet mechanism, and 
the original Sequicity model. It is immediately apparent that the Copynet improves both
the F1 and BLEU scores of our model. 
The Transformer with Copynet also learns to produce responses more similar 
to the gold target sequences 
at a cost of lower success rate compared to the original Sequicity model.

\subsection{Example outputs}
These are some examples of the system outputs. 
Overall, the outputs of our Transformer + copynet model look comparable to Sequicity. The model learns correct behavior for the domain, but chooses responses that contain less slot placeholders than the human reference, which reflects in the lower Success F1.

In the first case, both models produce reasonable responses. Our model is closer to the reference in both semantic and n-gram overlap, but this is not reflected in the Success F1 score as there are no slot placeholders present.
In the second case, both models give very similar responses, which also overlap with the reference. In the third case, our model gives a response that paraphrases the reference and is more specific than the one of Sequicity, but, again, does not contain any slot placeholders, so Success F1 score is unable to capture it.

\begin{enumerate}
\item \begin{description}
\item[User] i want afghan food in the centre of town .
\item[Transformer (no copynet)] name\_SLOT is a nice restaurant in the city centre of town . 
\item[Transformer (copynet)] there are several restaurants in the centre of town . do you have a preference for ?
\item[Sequicity] there are no restaurants serving centre food .
\item[Real response] what price range are you looking for ?
\end{description}

\item \begin{description}
\item[User] yes , what is the address and phone number for pizza hut cherry hinton ?
\item[Transformer (no copynet)] name\_SLOT is in the area\_SLOT part of town and is in the pricerange\_SLOT price range .
\item[Transformer (copynet)] name\_SLOT is located at address\_SLOT and their phone number is phone\_SLOT . is there anything else i can help you with ?
\item[Sequicity] name\_SLOT is located at address\_SLOT and their phone number is phone\_SLOT . is there anything else i can help you with ? 
\item[Real response] the address for name\_SLOT is address\_SLOT and the phone number is phone\_SLOT 
\end{description}

\item \begin{description}
\item[User] i want a restaurant in the north part of town .
\item[Transformer (no copynet)] name\_SLOT is a moderately priced restaurant in the north part of town .
\item[Transformer (copynet)] there are two restaurants in the north part of town . do you want the address ?
\item[Sequicity] there are several restaurants in the north part of town . 
\item[Real response] the name\_SLOT is in the north part of town . the name\_SLOT is also located in the north part of town .
\end{description}

\end{enumerate}

\section{Conclusion}
In this paper, we have investigated the use of Transformer as the model for end-to-end dialogue system.
We have shown that it has comparable performance with Sequicity.
We believe the system could be improved by utilizing a pre-trained language model \cite{devlin, radford, yang, song}.
Although the the success F1 score did not supersede our baseline, our model has BLEU score of responses $7.4\%$ higher than Sequicity. 


\newpage
\bibliographystyle{apalike}
\bibliography{references}

\end{document}